\documentclass{article}

\PassOptionsToPackage{numbers}{natbib}
\usepackage[final]{neurips_2022}

\usepackage[utf8]{inputenc} 
\usepackage[T1]{fontenc}    
\usepackage{hyperref}       
\usepackage{url}            
\usepackage{booktabs}       
\usepackage{amsfonts}       
\usepackage{nicefrac}       
\usepackage{microtype}      
\usepackage{xcolor}         

\usepackage{graphicx}
\usepackage{import}
\usepackage{pdfpages}
\usepackage{transparent}
\usepackage{svg}
\usepackage[english]{babel}
\usepackage{amsmath}
\usepackage{xfrac}
\usepackage{caption}
\usepackage{subcaption}

\title{Simulation-Based Parallel Training}

\author{%
    Lucas ~Meyer \\
    Industrial AI Laboratory SINCLAIR, EDF Lab Paris-Saclay, \\
    Univ. Grenoble Alpes, Inria, CNRS, Grenoble INP, LIG \\
    \texttt{lucas.meyer@inria.fr} \\
    \And
    Alejandro ~Ribes \\
    Industrial AI Laboratory SINCLAIR, EDF Lab Paris-Saclay \\
    91120 Palaiseau, France \\
    \texttt{alejandro.ribes@edf.fr} \\
    \And
    Bruno ~Raffin \\
    Univ. Grenoble Alpes, Inria, CNRS, Grenoble INP, LIG \\
    38000 Grenoble, France \\
    \texttt{bruno.raffin@inria.fr} \\
}
\date{September 2022}

\begin{document}

\maketitle

\begin{abstract}

    Numerical simulations are ubiquitous in science and engineering.  Machine
    learning for science investigates how artificial neural architectures can
    learn from these simulations to speed up scientific discovery and
    engineering processes. Most of these architectures are trained in a
    supervised manner. They require tremendous amounts of data from simulations
    that are slow to generate and memory greedy. In this article, we present our
    ongoing work to design a training framework that alleviates those
    bottlenecks. It generates data in parallel with the training process.  Such
    simultaneity induces a bias in the data available during the training.  We
    present a strategy to mitigate this bias with a memory buffer. We test our
    framework on the multi-parametric Lorenz's attractor. We show the benefit of
    our framework compared to offline training and the success of our data bias
    mitigation strategy to capture the complex chaotic dynamics of the system.

\end{abstract}

\section[]{Introduction}

Multiparametric dynamical simulations are crucial to scientific discovery and
engineering applications from computational fluid dynamics \cite{moin1998direct}
to biochemistry \cite{abraham2015gromacs}. Let's denote $g$ such a simulation,
$X$ its input parameters and $Y^t_X$ the output for a given time step $t$. 

\begin{align*}
g\colon \mathbb{R}^{d_{\text{in}}} &\to \mathbb{R}^{d_{\text{out}} \times T} \\
X & \mapsto \{Y^t_X, t \in [0, T], T \in \mathbb{R}^+\}
\end{align*}

Recently, scientists have considered deep learning as a means to accelerate
science by speeding up those simulations and learning from them
\cite{stevens2020ai}. Most works train an artificial neural network $f_\theta$
in a supervised manner, from data generated by the simulation process $g$. The
exact combination of inputs and outputs of the training process depends on the
application. For instance, for a network trained to replace the original
simulator $g$ with direct time predictions, as done in the seminal work of
Raissi et al.  \cite{raissi2019physics}, we have $f_\theta(X, t) \approx Y^t_X$.
However, for the auto-regressive models proposed in
\cite{pfaff2020learning,brandstetter2021message}, we have $f_\theta(X,
Y^{t}_X) \approx Y^{t+1}_X$.

Classical training consists in generating a set of simulation data
$\mathcal{X}_{\text{train}}$, $\mathcal{Y}_{\text{train}}$. Where
$\mathcal{X}_{\text{train}}$ would be representative of a parameter set on which
we want to infer with the trained neural network. Efficient training $f_\theta$,
and capturing all the complexity of the simulated system, generally requires
tremendous amounts of data \cite{brunton2020machine}. The dimension of
$\mathcal{Y}_{\text{train}}$ can thus quickly explode and lead to a massive
memory footprint. Additionally, the sequential generation of
$\mathcal{Y}_{\text{train}}$ can be prohibitively slow. To overcome these
limitations, one may restrict $\mathcal{X}_{\text{train}}$, or subsample
$\mathcal{Y}_{\text{train}}$ by reducing the number of time steps or observed
dimensions. To reduce $\mathcal{X}_{\text{train}}$ does not necessarily hinder
the quality of the training. Techniques ranging from adaptive experimental design
\cite{foster2021deep} to importance sampling \cite{katharopoulos2018not} help to
identify the parameters that will bring the most relevant information for the
training of the network. These techniques, however, require knowledge about the
network response to different inputs $X$, which is not feasible if
$\mathcal{X}_{\text{train}}$ is set beforehand. 

We propose a framework that generates the simulation data simultaneously with
the training process. Data generation is highly parallelized and streamed to
training to avoid storing any data in files. The framework also allows the
steering of the training by selecting new parameters $X$ based on some feedback
from the network on previously presented parameters.  This article presents our
ongoing work covering the first point. Compared to classical training from
files, the simultaneity of the training and the data generation presents
specificities that must be addressed.

At any time $\tau$ of the training process, only the already generated data
$\mathcal{Y}^\tau$ can be presented to the network. There is no guarantee that
$\mathcal{Y}^\tau$ is representative of $\mathcal{Y}_{\text{train}}$. Because of
the limited number of concurrent simulations $g$ and their iterative production
of data, the two sets can have very distinct distributions. This setting
commonly arises in online deep learning \cite{ditzler2015learning}. If we stream
the data, using the last received time step $Y^\tau$ directly for training, this
is known to lead to poor performances related to catastrophic forgetting
\cite{kemker2018measuring}. Therefore, to mitigate the bias in the data due to
the discrepancies between $\mathcal{Y}_{\text{train}}$ and $\mathcal{Y}^\tau$,
we introduce a \emph{memory buffer} of limited size. It is associated with a
\emph{data management policy} in charge of the selection of elements to
construct batches and their eviction when the buffer is full.  Moreover, we rely
on a simple sampling strategy of $\mathcal{X}_{\text{train}}$ to improve the
diversity of $\mathcal{Y}^\tau$. 

We validate our framework and the strategy to mitigate data bias on a model that
captures the chaotic dynamics of Lorenz's attractor
\cite{lorenz1963deterministic}. First, we show that our data bias mitigation
strategy is successful in producing more diverse batches than pure streaming
learning. Second, we show that our framework can prevent the overfitting offline
training suffers, by exposing the network to more diverse trajectories. This
overfitting typically occurs when the simulation data are too big to be stored,
which imposes to limit the size of the dataset and consequently reduces its
representativeness. Third, we show that the framework can match the performances
of offline training with a comprehensive dataset. This illustrates there is no
performance decrease using the online framework.

The main contributions of our ongoing work are:
\begin{itemize}

    \item{a deep learning framework orchestrating parallel data simulations with
    the online  training of neural architectures on multi-parametric dynamical
    systems;}

    \item{a strategy to mitigate the bias in data induced by the online nature
    of the framework.}

\end{itemize}

\section{Related Work}

\subsection{Deep Learning and Numerical Simulations}

In recent years, deep learning has permeated the numerical simulations of
multi-parametric dynamical systems. Scientists proposed algorithms to substitute
the traditional partial differential equations (PDE) solvers
\cite{raissi2019physics,li2020fourier,brandstetter2021message}. Instead of
replacing entirely the original solver, others proposed to augment the
resolution of coarse solutions \cite{um2020solver,kochkov2021machine}.
Similarly, Erichson et al. presented a network to reconstruct the dynamics of a
fluid from a few data points only \cite{erichson2020shallow}. Others have relied
on numerical simulations to study how machine learning interfaces with
traditional dynamical system theories, like the Koopman operator
\cite{lusch2018deep}, or help to rediscover the governing equations of simulated
systems \cite{cranmer2020discovering}.

All these approaches rely on supervised training, for which data mostly come
from simulators. Even approaches that could theoretically be data-free, like
physics-informed neural networks \cite{raissi2019physics}, have required
simulated data to work successfully on complex physics \cite{lucor2022simple}.

Most of the published work is only presented with fairly simple simulations,
either in terms of the complexity of the governing equations or by the scale of
the problem at stake. Even though some works propose to capture intrinsic
properties of the studied physical systems, it is, to our knowledge, not yet
clear how deep learning architectures trained on simple problems can scale to
larger and more complex ones, common to the industry and today's scientific
challenges. For instance, using graph neural networks, Pfaff et al. could infer
dynamics with higher resolution than the one used for training
\cite{pfaff2020learning}. Equivariance is another example to learn fundamental
properties of a system, like symmetries, and thus reduce the amount of data
required for training \cite{satorras2021n}. 

\subsection{Online Training}

By generating the simulations along the training process the full dataset is not
accessible anymore at all times. In this setting referred to as online learning
\cite{ditzler2015learning}, data presented to the network may be biased. It
leads to poor generalization performances, like catastrophic forgetting
\cite{kemker2018measuring}. Several techniques have been proposed to mitigate
this catastrophic forgetting. Either by tuning the loss for the optimal weights
of the network not to change dramatically while being trained on new data 
\cite{kirkpatrick2017overcoming,lee2017overcoming}. Another approach consists in
repeating the samples previously shown to the model \cite{hayes2019memory}. Such
a strategy has also been successfully applied in the context of deep
reinforcement learning and coined as replay buffer by the community
\cite{schaul2015prioritized,rolnick2019experience}.

\subsection{Parallel Frameworks for Deep Reinforcement Learning}

Training frameworks presenting several degrees of parallelism have already been
proposed by the deep reinforcement learning community
\cite{nair2015massively,mnih2016asynchronous,horgan2018distributed}. Especially,
to achieve the acting and the learning in parallel.  Similarly, in our
framework, the data generation and the training are parallelized. Liang et al.
propose, within the Ray project, a task-based formalism to abstract those
different degrees of parallelism for deep reinforcement learning
\cite{liang2018rllib}.  If some reinforcement learning experiments lead to
massive parallel resource usage due to the very large number of actors to run,
the simulation itself is not massively parallelized \cite{berner2019dota}.
Indeed, these deep reinforcement learning framework support simulations that are
either sequential or multi-threaded, but not scientific PDEs solvers requiring
distributed memory parallelization with  MPI to address problems at a relevant
scale.

Despite the similarity in the parallelism of the data generation and the
training, the comparison with deep reinforcement learning has limits.  We are
not learning to interact with an environment but rather learning the environment
itself.  Therefore there is no policy, and thus no concept of off-policy. All
samples are relevant, although some may bear more useful information for the
training of the network. It is also worth mentioning the work of Brace et al.,
which introduces a parallel framework to select biomolecular simulations to run,
but not to train generically architectures on numerical simulations
\cite{brace2022coupling}.

\section[]{Training Framework}

\subsection{Architecture Overview}


\begin{figure}[ht!]
    \centering 
    \includegraphics[width=0.8\textwidth]{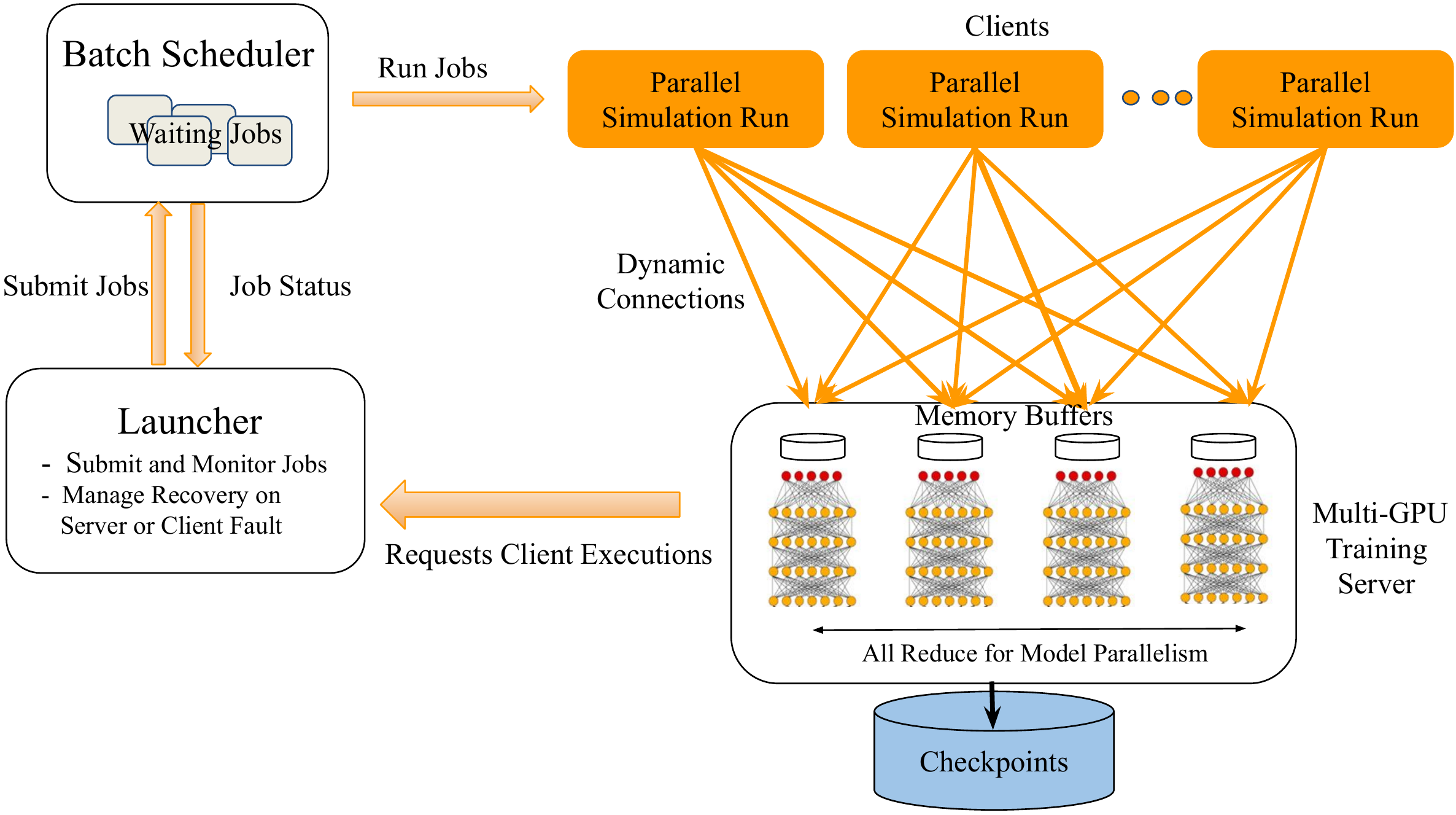}

    \caption{Online Training Framework Overview. The launcher starts and
    monitors the clients and the server. The server is in charge of the training
    loop, which can be executed in parallel on several GPUs. It controls the
    experimental design and requests to the launcher the executions of new
    client instances for selected sets of parameters.  The launcher can start
    these instances once the batch scheduler has allocated the required compute
    resources.}

    \label{fig:deepmelissa-arch} 

\end{figure}

Our framework\footnote{\url{https://gitlab.inria.fr/melissa}} is based
on an adaptation of the open source Melissa architecture developed to manage
large-scale ensemble runs for sensibility analysis~\cite{terraz2017melissa} and
data assimilation~\cite{friedemann-melissaDA:2022}. Our goal is to enable the
online (no intermediate storage in files)  training of a neural model from data
generated through different simulation executions. These simulations are
executed with different input parameters (parameter sweep), making the different
members of the ensemble.
 
 We target executions on supercomputers where simulations can be large parallel
 solver codes executed on several nodes and the training parallelized on several
 GPUs (Figure \ref{fig:deepmelissa-arch}). We give below an overview of the
 framework, a detailed description being beyond the scope of this paper. The
 framework relies on three main components:

  \begin{itemize}

  \item{The launcher orchestrates the execution of the simulation instances and
  the training. The launcher interacts with the machine batch
  scheduler (e.g. Slurm or OAR) to request resources for the other
  component executions.}

  \item{Each client executes one simulation instance. When starting, it
  dynamically connects to the server. As soon as it produces new data (a new
  time step), the data is sent to the server.} 

  \item{The server runs on several GPUs to train the model in parallel. On
  reception of new data from one client, the data is first pushed into the local
  memory buffer. When a new batch is required for training, it is extracted from
  this buffer. The management of this buffer is detailed in the following
  section \ref{sec:memory_buffer}.}

  \end{itemize}

 This framework thus alleviates the need for files to store intermediate
 results. We target large-scale executions generating large amounts of data that
 cannot be reasonably stored. As an example, Melissa has been reported
 performing a sensibility analysis run executing $8 000$ simulation instances
 that generated $48$TB of data processed online \cite{terraz2017melissa}. The
 secondary benefit of coupling data generation and training is that the server
 can control which simulation instances to run next, opening the possibility for
 adaptive experimental design.

 The framework combines several levels of parallelism (at the ensemble, client,
 server, and communication levels) to enable efficient and scalable executions.
 The simulation can be a parallel code written in C/C++, Fortran or Python, 
 and combining the usage of MPI, OpenMP and Cuda.  The server relies on Pytorch Data
 Distributed Parallelism \cite{li2020pytorch} for enabling multi-GPU training.

\subsection{Data Bias Mitigation Strategy}
\label{sec:memory_buffer}

\begin{figure}[ht!]
    \centering
    \includegraphics[width=0.8\textwidth]{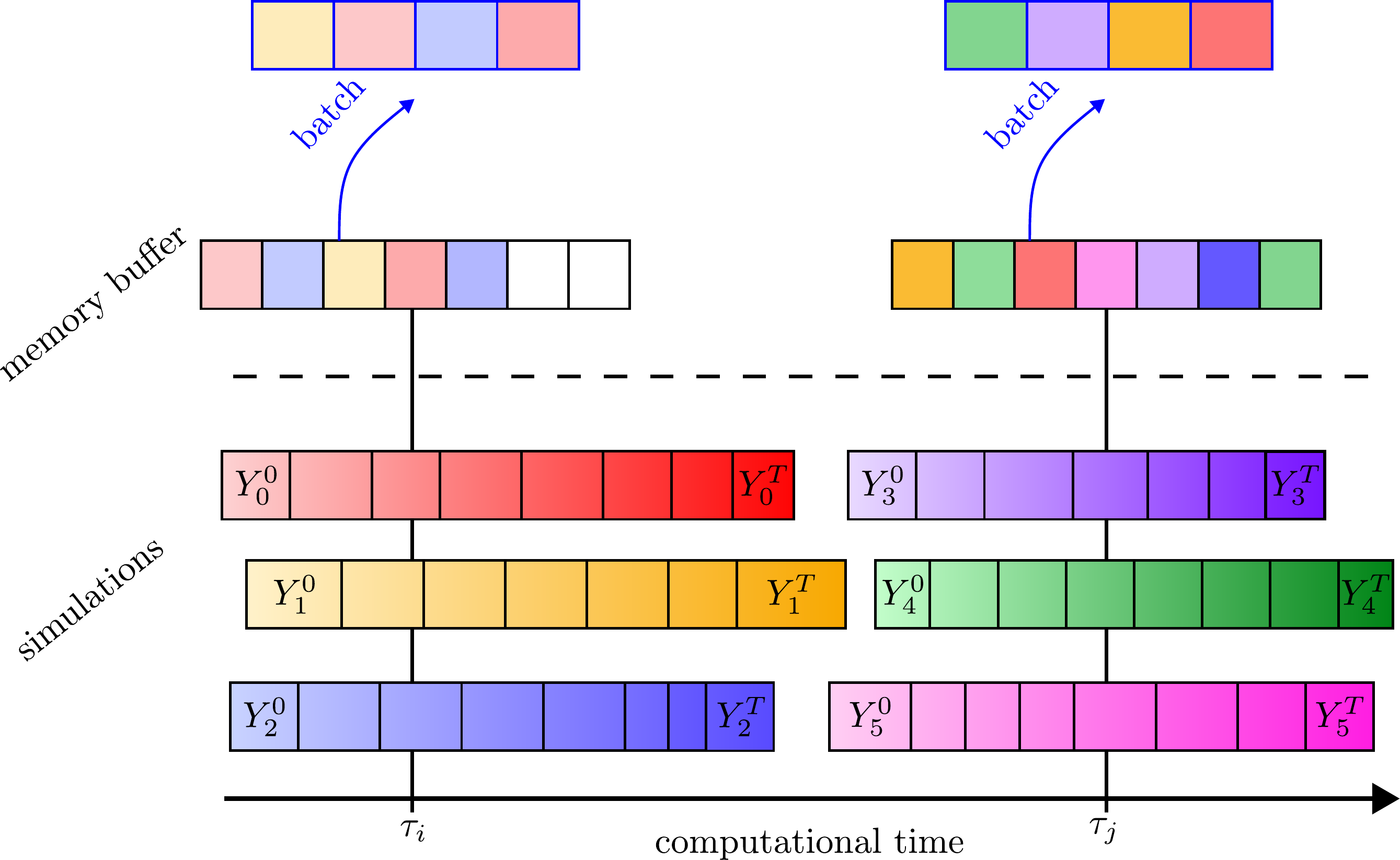}

    \caption{Timeline of the generated simulation data. Here, 3 processes run 6
    simulations in parallel. Groups of concurrent simulations are executed
    sequentially. As soon as a simulation time step $Y^t_{X_i}$ is computed, it is
    passed to the memory buffer. Simultaneously, this cache of limited capacity
    yields batches from previously stored time steps. The content of the buffer
    and the batches constructed from it are represented at two different times
    $\tau_i$ and $\tau_j$. At $\tau_i$ the buffer is not yet full. At $\tau_j$
    the buffer is full and contains some time steps from previous simulations.} 

    \label{fig:data-transfer}
\end{figure}

 The online data generation inherently carries 3 types of data bias
 (Figure \ref{fig:data-transfer}):

 \begin{itemize}

 \item{Intra-simulation bias. Simulators produce dynamics as time series by
 discretizing time and progressing iteratively. After $T$ time steps, only the data
 $\{Y^t_{X}, 0 \leq t \leq T \}$ for the simulation instance $X$ are available
 for training.}

\item{Inter-simulation bias. Computational resources are limited and so often not
all simulation instances can be executed concurrently. Let's assume that only
$c$ simulations can execute concurrently at any time.  At the end of the
execution of the $c$ first simulations training only has access to the data
$\{Y^t_{X_i}, 0 < i \leq c \}$. }

\item{Memory bias. Not only online training cannot have access to the {\it not
already generated data}, but we also cannot keep all {\it already generated
data} due to memory constraints. We can, at most, keep a rolling sample of data 
that fits into a given memory budget.} 

\end{itemize}

Therefore, batches cannot be drawn uniformly from the full dataset as performed
with traditional epoch-based training.  This bias has a detrimental effect on
the training, partly due to catastrophic forgetting \cite{kemker2018measuring}.

Some physical systems exhibit different behaviors depending on the parameter of
the simulation. For instance, Reynolds number for fluid dynamics
\cite{ferziger2002computational}.  The inter-simulation bias creates
discrepancies between $\mathcal{Y}^{\tau}$ and $\mathcal{Y}_{\text{train}}$. The
\emph{sampling strategy} consists in selecting parameter $X$ so that
$\mathcal{Y}^{\tau}$ is as representative as possible of all the behaviors in
$\mathcal{Y}_{\text{train}}$. In the following, we adopt a simple strategy that
randomly samples $X$ among $\mathcal{X}_{\text{train}}$.

To mix the data further and address the intra-simulation and memory biases, we
introduce a \emph{memory buffer}. It acts like a cache of the data previously
generated and sent to each trainer. Without the memory buffer, data would be
immediately streamed from the simulators to the trainers, a situation prone to
catastrophic forgetting. The memory buffer has a set capacity, $s$. It is
associated with a \emph{data management policy} that specifies the rule to evict
data when the buffer is full, and select data to construct the batches. The
current policy selects samples randomly from the buffer and
erases them upon selection. This avoids having samples repeated too many times
in batches while the trainer is waiting for new data from the simulators.  Due
to the intra-simulation bias, the first simulation time steps arrive first in
the memory buffer. To avoid yielding first batches with overly represented first
time steps, we set a threshold $s_\text{ready}$ that signals when the buffer has
enough elements to be sampled. Because the training loop and the simulation are
executed at different speeds on different processes, the memory buffer must also
amortize the data production and consumption between the clients and the
trainer. We set a second threshold $s_\text{min}$ that ensures there is always a
minimal amount of samples in the buffer, for the batch construction to be indeed
random. 

In summary, the key idea is that the simultaneous generation of simulation data
creates a bias in the data. We rely on adequate sampling of the simulation
parameter and a memory buffer to mix the data presented to the network and
guarantee the training quality.

\section[]{Experiments}

To prove the usefulness of our approach we examine the case of a neural network
trained to capture the underlying physics from the observation of complex
dynamics. For this purpose, we consider Lorenz's attractor
\cite{lorenz1963deterministic}. Its dynamics are described by the system
\ref{eq:lorenz}. The attractor is chaotic. It means that, despite being
deterministic, two points initially close have very distinct trajectories. The
dynamics thus present a complex and rich behavior.  Nonetheless, the underlying
non-linear equation is fairly simple. The possibility to recover the dynamics of
Lorenz using machine learning techniques has already been studied in
\cite{brunton2016discovering,dubois2020data}, even with partial observation
\cite{lin2021data}. Here, we train a neural network to reconstruct the
trajectory with an integration scheme of Euler, i.e. we predict $f_\theta(X,
Y_t) \approx \frac{Y_{t+1} - Y_t}{\Delta t}$, where $X = \rho$ and $Y_t =
(x_t, y_t, z_t)$ the coordinates of the system at time $t$ given by Equation
\ref{eq:lorenz}. The model is auto-regressive as the trajectory is reconstructed
iteratively at inference. 

\begin{align}
   \frac{dx}{dt} &= \sigma(y-z) \nonumber \\
   \frac{dy}{dt} &= x(\rho - z) - y \nonumber \\
   \frac{dz}{dt} &= xy - \beta z \nonumber \\
   \label{eq:lorenz}
\end{align}

We consider the multi-parametric case for which the initial position $X_0$ and
the parameter $\rho$ vary.  $\sigma$ and $\beta$ are respectively fixed to $10$
and $\sfrac{8}{3}$. The goal is to train a neural network that generalizes for
any initial position and $\rho$ belonging to the interval $[0, 100]$. In the
following experiments, initial positions $X_0$ are always taken from the normal
distribution $\mathcal{N}(15, 30)$. Data are standardized with statistics
computed beforehand. The model architecture also remains the same across the
experiments. It consists of 3 layers of 512 features. The first layer has 4
input features for the position and the parameter $\rho$. The output has 3
features for the velocity. We use the SiLU activation function between each
intermediate layer \cite{hendrycks2016gaussian}. For the memory buffer, we
arbitrarily set $s$, $s_\text{ready}$, and $s_\text{min}$ equivalent to 10, 5
and 3 simulations respectively.

In our experiments, we consider different datasets and training settings. We
always generate a different number of trajectories for values of $\rho$ taken in
$[0, 20, 40, 60, 80, 100]$. First, we have a \emph{full offline} dataset
consisting of 100 trajectories of 2000 time steps $\Delta t$ of $10^{-2}$
second. A second offline dataset, we call \emph{restricted}, consists of only 10
trajectories. A third offline dataset, referred to as \emph{subsampled},
corresponds to 10000 trajectories sampled every 100 time steps. The restricted
and subsampled offline datasets are artificially reduced to illustrate the case
of more complex simulations for which it might be difficult to obtain and store
a representative dataset. As we compare different dataset sizes and different
training strategies (offline or online), for a fair comparison, we ensure that
the quantity of information seen by the network is always the same. This
translates into a constant total number of batches. For online training, there
is no notion of epoch anymore. Hence, the training of the full, restricted, and
subsampled datasets is respectively achieved on 100, 1000, and 100 epochs. For
the online dataset, we generate 10000 trajectories. In the \emph{streaming}
setting, the trajectories are generated from simulations with increasing values
of $\rho$.  Data are directly gathered into batches and passed to the trainer. A
second online setting, noted as \emph{sampling}, randomly samples $\rho$ prior
to the data generation. The online setting, noted as \emph{sampling + memory
buffer}, combines the random sampling of the parameters and a memory buffer with
random data selection policy. For all training strategies, the batch size is
fixed to 1024 time steps.  The validation dataset consists of 10 trajectories, 2
for each value of $\rho$ taken in the same interval as for the training set.

\subsection{Memory Buffer Mitigates the Bias Induced by Online Training}

\begin{figure}[ht!]
    \centering
    \includegraphics[width=.7\textwidth]{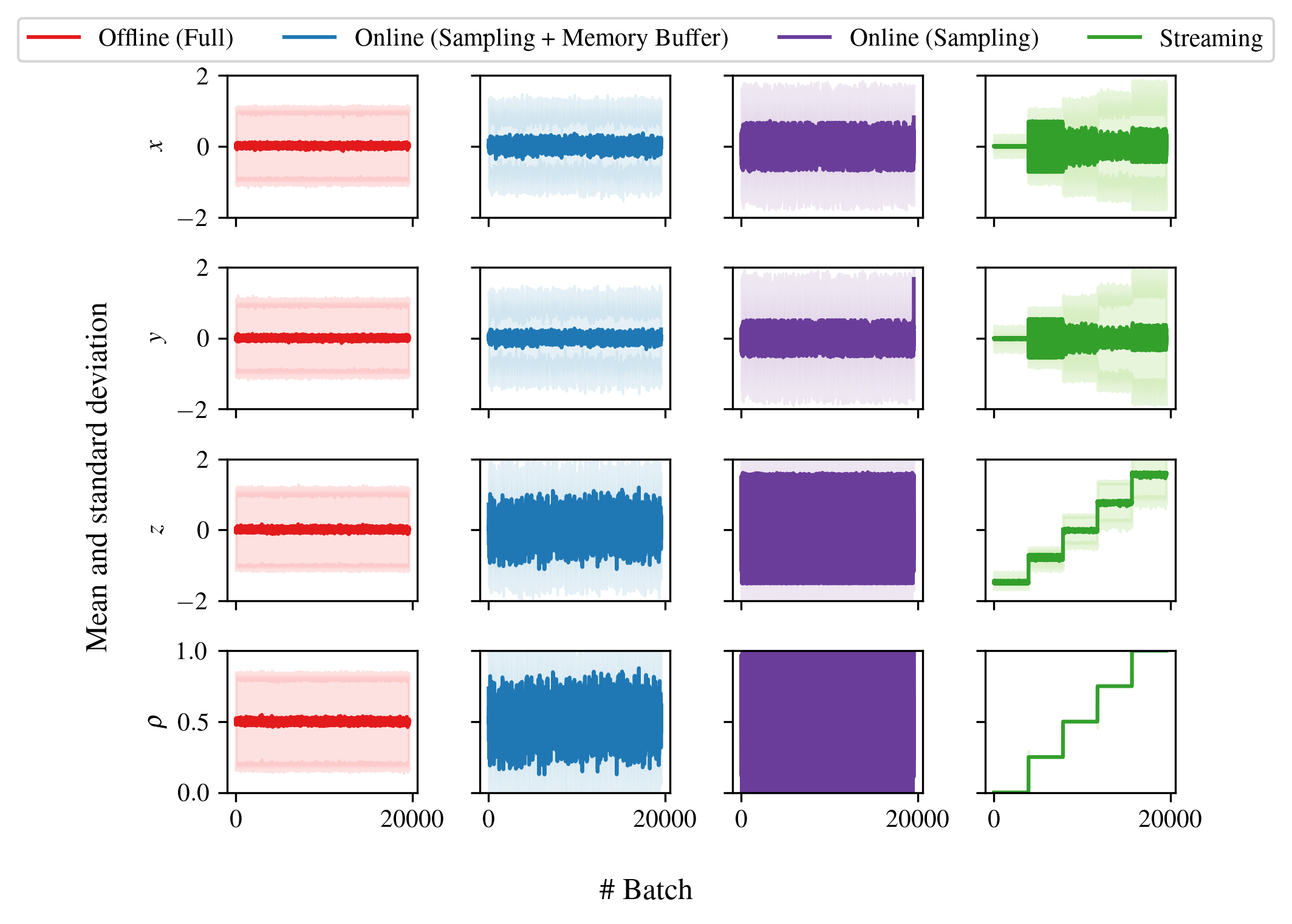}
    
    \caption{Comparison of batch statistics for different training strategies.
    Offline training presents batches uniformly drawn from the dataset, which is
    assumed to lead to the best performances. Sampling and memory buffer correct
    the prejudicial bias observed in streaming batches.}
    
    \label{fig:batch-statistics}
\end{figure}

To illustrate the efficiency of the two levels of control, namely the sampling
of parameters and the memory buffer, in mitigating the data bias,  we consider
the batch statistics in the different training settings. Figure
\ref{fig:batch-statistics} shows the mean and standard deviation of the batches
for the different strategies. The streaming strategy appears biased, exposing
the training process to catastrophic forgetting that can explain the gap between
train and validation losses in Figure \ref{fig:training-strategies}.  We see
that the coupling of the sampling strategy and the memory buffer makes the
distribution closer to the one observed in the offline setting, showing its
capability to reduce bias.

\subsection{Parameter Space Exploration with Online Training} 
\label{sec:space_exploration}

Training and validation losses for the different training settings are displayed
in Figure \ref{fig:training-strategies}. We observe that for the restricted and
the subsampled offline datasets the validation losses are almost an order higher
than the training losses, which indicates overfitting. This shows that, even for
simple equations like Lorenz's system (Equation \ref{eq:lorenz}), a neural network can
experience difficulties in recovering the dynamics with only a few trajectories.
It stresses the importance of having a representative dataset. For the same
number of batches, the online training, with a random sampling strategy and the
memory buffer, presents training and validation losses that decrease similarly
with the number of batches. It is also for this setting that the validation loss
is minimal. By presenting to the network more diverse trajectories, this
strategy overcomes the overfitting observed with limited offline datasets.

\begin{figure}[ht!]
    \centering
    \includegraphics[width=.7\textwidth]{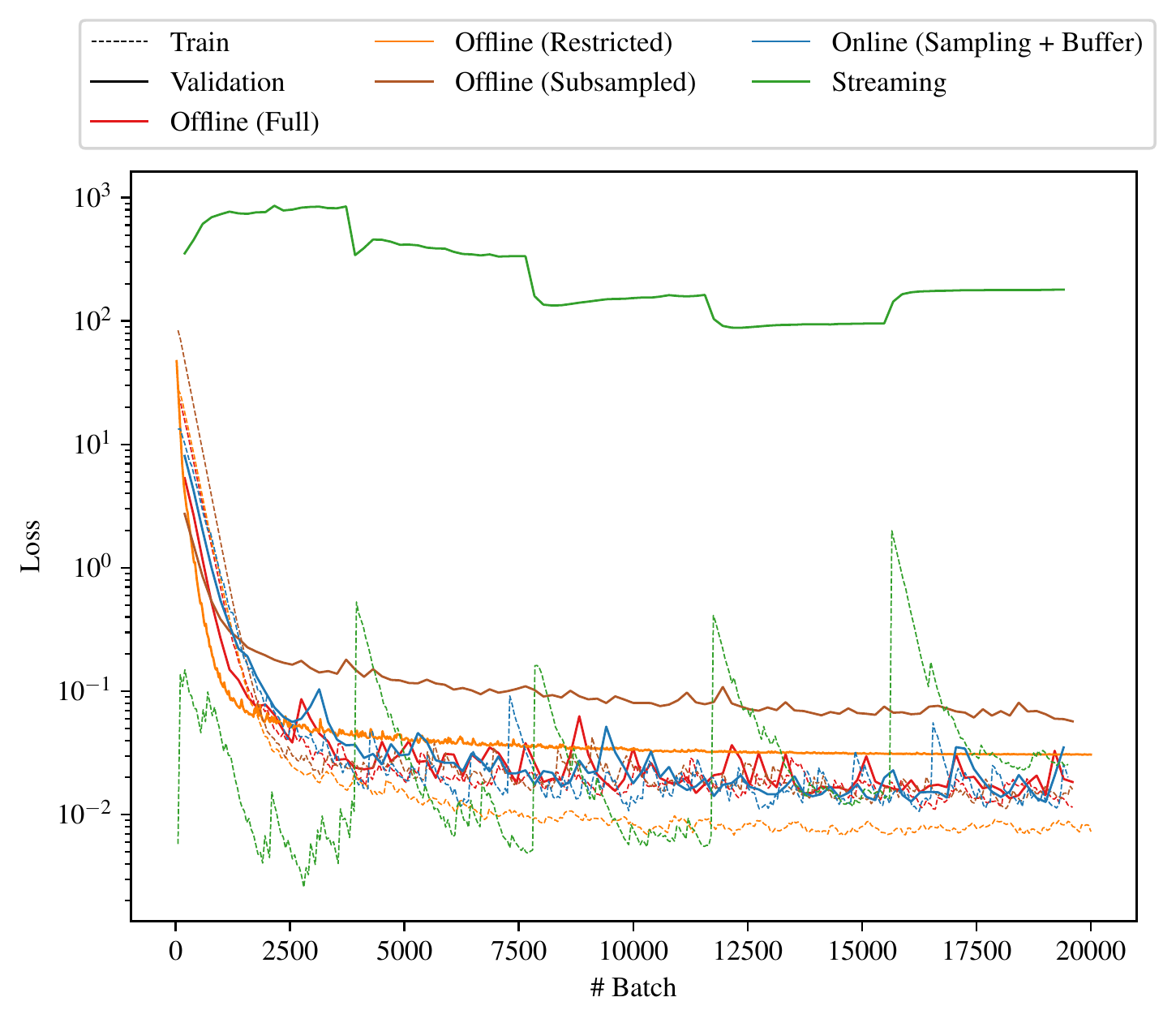}

    \caption{Training and validation in plain and dashed lines respectively for
    different training strategies.} 

    \label{fig:training-strategiester}
\end{figure}

\subsection{Recovery of Offline Performances}

\begin{figure}[ht!]
    \centering    
    \begin{subfigure}[b]{.45\textwidth}
        \includegraphics[width=\textwidth]{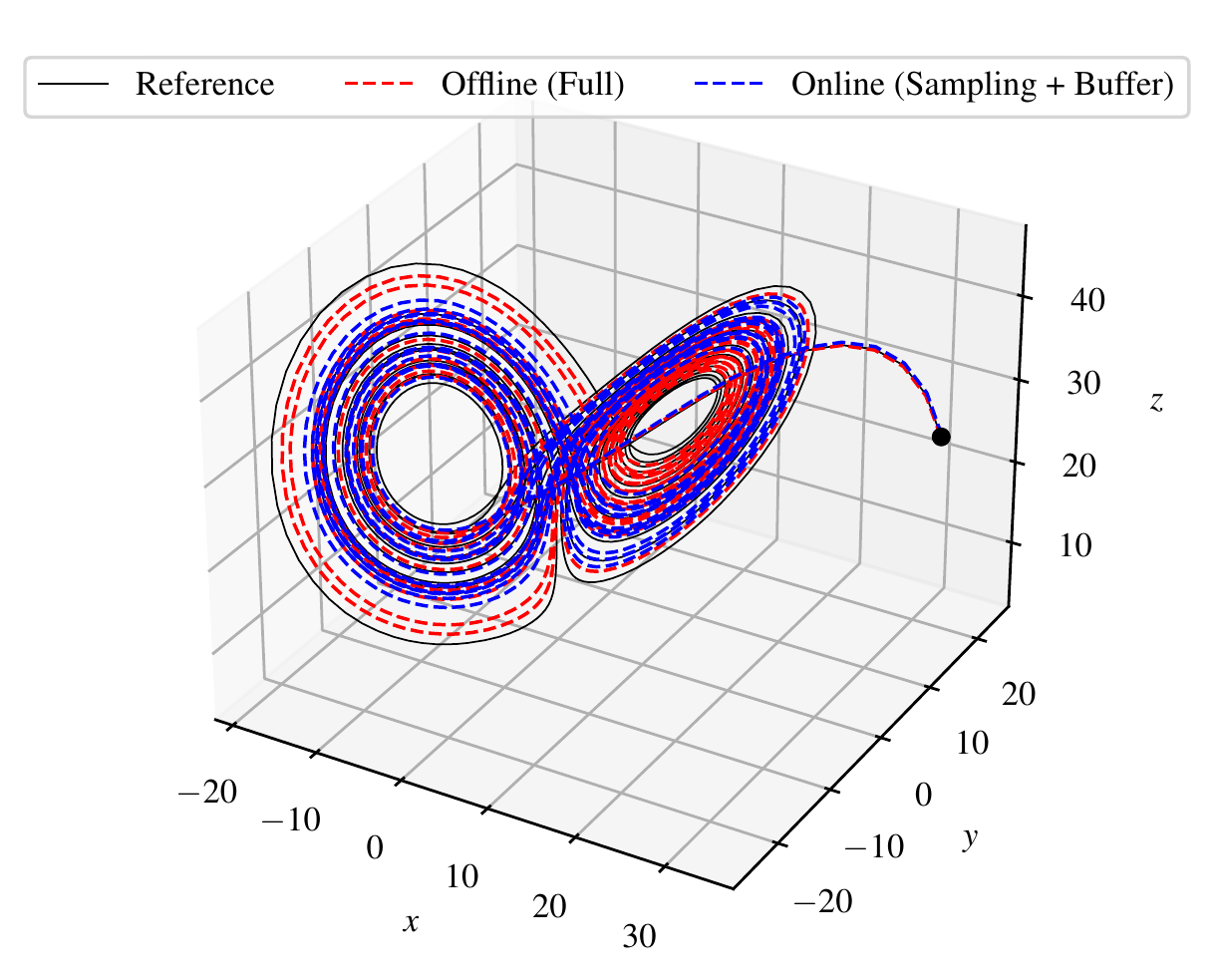}
        \caption{3D trajectory}
        \label{fig:3D-traj}
    \end{subfigure}
    \hfill
    \begin{subfigure}[b]{.45\textwidth}
        \includegraphics[width=\textwidth]{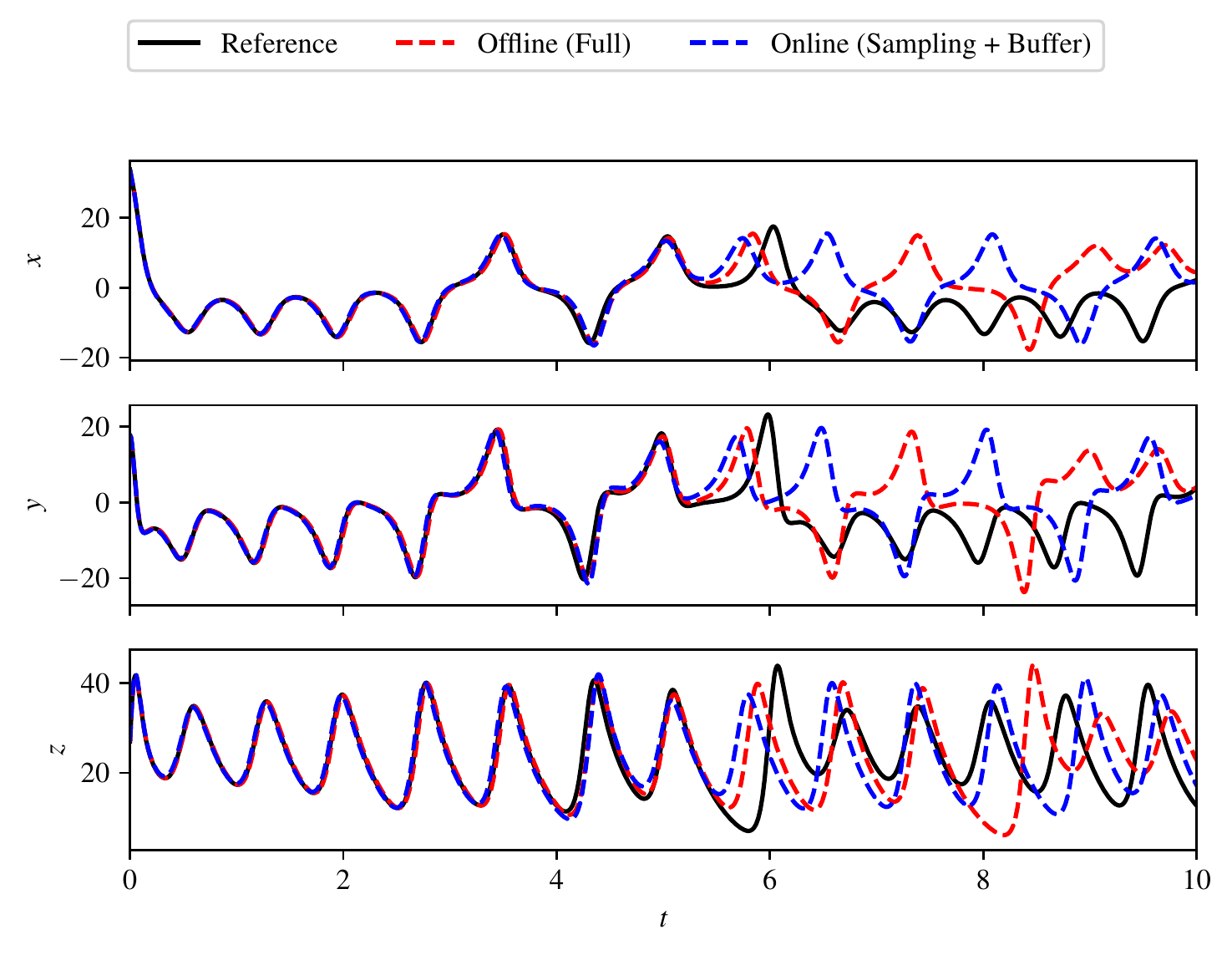}
        \caption{Per axis trajectory}
        \label{fig:axis-traj}
    \end{subfigure}

    \caption{Comparison of the model predictions on a test trajectory for
    different training settings. Both settings manage to train the network to
    accurately capture the dynamics of Lorenz's attractor.}

    \label{fig:trajectories}
\end{figure}

Figure \ref{fig:training-strategies} shows the training dynamics with the full
offline dataset and the online setting (sampling + buffer) are equivalent.  Both
configurations correctly recover the dynamics of the attractor on a test
trajectory generated with $\rho$ equal to 28 (Figure \ref{fig:trajectories}).
The predicted trajectories are stable as shown in Figure \ref{fig:3D-traj}. They
match the reference for a limited time, as seen in Figure \ref{fig:axis-traj}.
The lack of coherence after some time is expected due to the chaotic nature of
Lorenz's attractor. Compared to offline training with a comprehensive dataset,
which can be difficult to obtain for more complex systems, the online strategy
does not induce any degradation of the trained network performances.

\section[]{Conclusion}

In this article, we have proposed a framework to train artificial neural
networks on synthetic simulations generated along the training process. We have
shown this setting induces a bias in the data that hinders the quality of the
training. We have introduced simple strategies to mitigate the bias: a random
sampling of the simulation parameters and a memory buffer that allows mixing
further the data presented to the network. We have applied these strategies to
train a network on Lorenz's attractor. We have shown that leveraging the online
setting to generate more data, but still keeping the same number of batches as
for an offline strategy, can lead to a lower validation loss. We have also shown
the strategies were successful to match the performance of offline training with
a comprehensive dataset in capturing the underlying dynamics of Lorenz's
atttractor, without incurring the same memory footprint. Such results need
further testing with different use cases to be confirmed. Advanced
implementations of the memory buffer will also be further investigated like
reservoir or importance sampling for the selection of elements in the memory
buffer \cite{vitter1985random,katharopoulos2018not}. 

\section*{Acknowledgements}

This project has received funding from the European High-Performance Computing Joint Undertaking (JU) under grant agreement No 956560.

\bibliography{references}

\begin{thebibliography}{40}
\providecommand{\natexlab}[1]{#1}
\providecommand{\url}[1]{\texttt{#1}}
\expandafter\ifx\csname urlstyle\endcsname\relax
  \providecommand{\doi}[1]{doi: #1}\else
  \providecommand{\doi}{doi: \begingroup \urlstyle{rm}\Url}\fi

\bibitem[Abraham et~al.(2015)Abraham, Murtola, Schulz, P{\'a}ll, Smith, Hess,
  and Lindahl]{abraham2015gromacs}
M.~J. Abraham, T.~Murtola, R.~Schulz, S.~P{\'a}ll, J.~C. Smith, B.~Hess, and
  E.~Lindahl.
\newblock Gromacs: High performance molecular simulations through multi-level
  parallelism from laptops to supercomputers.
\newblock \emph{SoftwareX}, 1:\penalty0 19--25, 2015.

\bibitem[Berner et~al.(2019)Berner, Brockman, Chan, Cheung, D{\k{e}}biak,
  Dennison, Farhi, Fischer, Hashme, Hesse, et~al.]{berner2019dota}
C.~Berner, G.~Brockman, B.~Chan, V.~Cheung, P.~D{\k{e}}biak, C.~Dennison,
  D.~Farhi, Q.~Fischer, S.~Hashme, C.~Hesse, et~al.
\newblock Dota 2 with large scale deep reinforcement learning.
\newblock \emph{arXiv preprint arXiv:1912.06680}, 2019.

\bibitem[Brace et~al.(2022)Brace, Yakushin, Ma, Trifan, Munson, Foster,
  Ramanathan, Lee, Turilli, and Jha]{brace2022coupling}
A.~Brace, I.~Yakushin, H.~Ma, A.~Trifan, T.~Munson, I.~Foster, A.~Ramanathan,
  H.~Lee, M.~Turilli, and S.~Jha.
\newblock Coupling streaming ai and hpc ensembles to achieve 100--1000$\times$
  faster biomolecular simulations.
\newblock In \emph{2022 IEEE International Parallel and Distributed Processing
  Symposium (IPDPS)}, pages 806--816. IEEE, 2022.

\bibitem[Brandstetter et~al.(2021)Brandstetter, Worrall, and
  Welling]{brandstetter2021message}
J.~Brandstetter, D.~E. Worrall, and M.~Welling.
\newblock Message passing neural pde solvers.
\newblock In \emph{International Conference on Learning Representations}, 2021.

\bibitem[Brunton et~al.(2016)Brunton, Proctor, and
  Kutz]{brunton2016discovering}
S.~L. Brunton, J.~L. Proctor, and J.~N. Kutz.
\newblock Discovering governing equations from data by sparse identification of
  nonlinear dynamical systems.
\newblock \emph{Proceedings of the national academy of sciences}, 113\penalty0
  (15):\penalty0 3932--3937, 2016.

\bibitem[Brunton et~al.(2020)Brunton, Noack, and
  Koumoutsakos]{brunton2020machine}
S.~L. Brunton, B.~R. Noack, and P.~Koumoutsakos.
\newblock Machine learning for fluid mechanics.
\newblock \emph{Annual review of fluid mechanics}, 52:\penalty0 477--508, 2020.

\bibitem[Cranmer et~al.(2020)Cranmer, Sanchez~Gonzalez, Battaglia, Xu, Cranmer,
  Spergel, and Ho]{cranmer2020discovering}
M.~Cranmer, A.~Sanchez~Gonzalez, P.~Battaglia, R.~Xu, K.~Cranmer, D.~Spergel,
  and S.~Ho.
\newblock Discovering symbolic models from deep learning with inductive biases.
\newblock \emph{Advances in Neural Information Processing Systems},
  33:\penalty0 17429--17442, 2020.

\bibitem[Ditzler et~al.(2015)Ditzler, Roveri, Alippi, and
  Polikar]{ditzler2015learning}
G.~Ditzler, M.~Roveri, C.~Alippi, and R.~Polikar.
\newblock Learning in nonstationary environments: A survey.
\newblock \emph{IEEE Computational Intelligence Magazine}, 10\penalty0
  (4):\penalty0 12--25, 2015.

\bibitem[Dubois et~al.(2020)Dubois, Gomez, Planckaert, and
  Perret]{dubois2020data}
P.~Dubois, T.~Gomez, L.~Planckaert, and L.~Perret.
\newblock Data-driven predictions of the lorenz system.
\newblock \emph{Physica D: Nonlinear Phenomena}, 408:\penalty0 132495, 2020.

\bibitem[Erichson et~al.(2020)Erichson, Mathelin, Yao, Brunton, Mahoney, and
  Kutz]{erichson2020shallow}
N.~B. Erichson, L.~Mathelin, Z.~Yao, S.~L. Brunton, M.~W. Mahoney, and J.~N.
  Kutz.
\newblock Shallow neural networks for fluid flow reconstruction with limited
  sensors.
\newblock \emph{Proceedings of the Royal Society A}, 476\penalty0
  (2238):\penalty0 20200097, 2020.

\bibitem[Ferziger et~al.(2002)Ferziger, Peri{\'c}, and
  Street]{ferziger2002computational}
J.~H. Ferziger, M.~Peri{\'c}, and R.~L. Street.
\newblock \emph{Computational methods for fluid dynamics}, volume~3.
\newblock Springer, 2002.

\bibitem[Foster et~al.(2021)Foster, Ivanova, Malik, and
  Rainforth]{foster2021deep}
A.~Foster, D.~R. Ivanova, I.~Malik, and T.~Rainforth.
\newblock Deep adaptive design: Amortizing sequential bayesian experimental
  design.
\newblock In \emph{International Conference on Machine Learning}, pages
  3384--3395. PMLR, 2021.

\bibitem[Friedemann and Raffin(2022)]{friedemann-melissaDA:2022}
S.~Friedemann and B.~Raffin.
\newblock {An elastic framework for ensemble-based large-scale data
  assimilation}.
\newblock \emph{{The international journal of high performance computing
  applications}}, 36\penalty0 (4):\penalty0 543--563, June 2022.
\newblock URL \url{https://hal.inria.fr/hal-03017033}.

\bibitem[Hayes et~al.(2019)Hayes, Cahill, and Kanan]{hayes2019memory}
T.~L. Hayes, N.~D. Cahill, and C.~Kanan.
\newblock Memory efficient experience replay for streaming learning.
\newblock In \emph{2019 International Conference on Robotics and Automation
  (ICRA)}, pages 9769--9776. IEEE, 2019.

\bibitem[Hendrycks and Gimpel(2016)]{hendrycks2016gaussian}
D.~Hendrycks and K.~Gimpel.
\newblock Gaussian error linear units (gelus).
\newblock \emph{arXiv preprint arXiv:1606.08415}, 2016.

\bibitem[Horgan et~al.(2018)Horgan, Quan, Budden, Barth-Maron, Hessel, van
  Hasselt, and Silver]{horgan2018distributed}
D.~Horgan, J.~Quan, D.~Budden, G.~Barth-Maron, M.~Hessel, H.~van Hasselt, and
  D.~Silver.
\newblock Distributed prioritized experience replay.
\newblock In \emph{International Conference on Learning Representations}, 2018.

\bibitem[Katharopoulos and Fleuret(2018)]{katharopoulos2018not}
A.~Katharopoulos and F.~Fleuret.
\newblock Not all samples are created equal: Deep learning with importance
  sampling.
\newblock In \emph{International conference on machine learning}, pages
  2525--2534. PMLR, 2018.

\bibitem[Kemker et~al.(2018)Kemker, McClure, Abitino, Hayes, and
  Kanan]{kemker2018measuring}
R.~Kemker, M.~McClure, A.~Abitino, T.~Hayes, and C.~Kanan.
\newblock Measuring catastrophic forgetting in neural networks.
\newblock In \emph{Proceedings of the AAAI Conference on Artificial
  Intelligence}, volume~32, 2018.

\bibitem[Kirkpatrick et~al.(2017)Kirkpatrick, Pascanu, Rabinowitz, Veness,
  Desjardins, Rusu, Milan, Quan, Ramalho, Grabska-Barwinska,
  et~al.]{kirkpatrick2017overcoming}
J.~Kirkpatrick, R.~Pascanu, N.~Rabinowitz, J.~Veness, G.~Desjardins, A.~A.
  Rusu, K.~Milan, J.~Quan, T.~Ramalho, A.~Grabska-Barwinska, et~al.
\newblock Overcoming catastrophic forgetting in neural networks.
\newblock \emph{Proceedings of the national academy of sciences}, 114\penalty0
  (13):\penalty0 3521--3526, 2017.

\bibitem[Kochkov et~al.(2021)Kochkov, Smith, Alieva, Wang, Brenner, and
  Hoyer]{kochkov2021machine}
D.~Kochkov, J.~A. Smith, A.~Alieva, Q.~Wang, M.~P. Brenner, and S.~Hoyer.
\newblock Machine learning--accelerated computational fluid dynamics.
\newblock \emph{Proceedings of the National Academy of Sciences}, 118\penalty0
  (21):\penalty0 e2101784118, 2021.

\bibitem[Lee et~al.(2017)Lee, Kim, Jun, Ha, and Zhang]{lee2017overcoming}
S.-W. Lee, J.-H. Kim, J.~Jun, J.-W. Ha, and B.-T. Zhang.
\newblock Overcoming catastrophic forgetting by incremental moment matching.
\newblock \emph{Advances in neural information processing systems}, 30, 2017.

\bibitem[Li et~al.(2020{\natexlab{a}})Li, Zhao, Varma, Salpekar, Noordhuis, Li,
  Paszke, Smith, Vaughan, Damania, et~al.]{li2020pytorch}
S.~Li, Y.~Zhao, R.~Varma, O.~Salpekar, P.~Noordhuis, T.~Li, A.~Paszke,
  J.~Smith, B.~Vaughan, P.~Damania, et~al.
\newblock Pytorch distributed: Experiences on accelerating data parallel
  training.
\newblock \emph{arXiv preprint arXiv:2006.15704}, 2020{\natexlab{a}}.

\bibitem[Li et~al.(2020{\natexlab{b}})Li, Kovachki, Azizzadenesheli,
  Bhattacharya, Stuart, Anandkumar, et~al.]{li2020fourier}
Z.~Li, N.~B. Kovachki, K.~Azizzadenesheli, K.~Bhattacharya, A.~Stuart,
  A.~Anandkumar, et~al.
\newblock Fourier neural operator for parametric partial differential
  equations.
\newblock In \emph{International Conference on Learning Representations},
  2020{\natexlab{b}}.

\bibitem[Liang et~al.(2018)Liang, Liaw, Nishihara, Moritz, Fox, Goldberg,
  Gonzalez, Jordan, and Stoica]{liang2018rllib}
E.~Liang, R.~Liaw, R.~Nishihara, P.~Moritz, R.~Fox, K.~Goldberg, J.~Gonzalez,
  M.~Jordan, and I.~Stoica.
\newblock Rllib: Abstractions for distributed reinforcement learning.
\newblock In \emph{International Conference on Machine Learning}, pages
  3053--3062. PMLR, 2018.

\bibitem[Lin et~al.(2021)Lin, Tian, Livescu, and Anghel]{lin2021data}
Y.~T. Lin, Y.~Tian, D.~Livescu, and M.~Anghel.
\newblock Data-driven learning for the mori--zwanzig formalism: A
  generalization of the koopman learning framework.
\newblock \emph{SIAM Journal on Applied Dynamical Systems}, 20\penalty0
  (4):\penalty0 2558--2601, 2021.

\bibitem[Lorenz(1963)]{lorenz1963deterministic}
E.~N. Lorenz.
\newblock Deterministic nonperiodic flow.
\newblock \emph{Journal of atmospheric sciences}, 20\penalty0 (2):\penalty0
  130--141, 1963.

\bibitem[Lucor et~al.(2022)Lucor, Agrawal, and Sergent]{lucor2022simple}
D.~Lucor, A.~Agrawal, and A.~Sergent.
\newblock Simple computational strategies for more effective physics-informed
  neural networks modeling of turbulent natural convection.
\newblock \emph{Journal of Computational Physics}, 456:\penalty0 111022, 2022.

\bibitem[Lusch et~al.(2018)Lusch, Kutz, and Brunton]{lusch2018deep}
B.~Lusch, J.~N. Kutz, and S.~L. Brunton.
\newblock Deep learning for universal linear embeddings of nonlinear dynamics.
\newblock \emph{Nature communications}, 9\penalty0 (1):\penalty0 1--10, 2018.

\bibitem[Mnih et~al.(2016)Mnih, Badia, Mirza, Graves, Lillicrap, Harley,
  Silver, and Kavukcuoglu]{mnih2016asynchronous}
V.~Mnih, A.~P. Badia, M.~Mirza, A.~Graves, T.~Lillicrap, T.~Harley, D.~Silver,
  and K.~Kavukcuoglu.
\newblock Asynchronous methods for deep reinforcement learning.
\newblock In \emph{International conference on machine learning}, pages
  1928--1937. PMLR, 2016.

\bibitem[Moin and Mahesh(1998)]{moin1998direct}
P.~Moin and K.~Mahesh.
\newblock Direct numerical simulation: a tool in turbulence research.
\newblock \emph{Annual review of fluid mechanics}, 30\penalty0 (1):\penalty0
  539--578, 1998.

\bibitem[Nair et~al.(2015)Nair, Srinivasan, Blackwell, Alcicek, Fearon,
  De~Maria, Panneershelvam, Suleyman, Beattie, Petersen,
  et~al.]{nair2015massively}
A.~Nair, P.~Srinivasan, S.~Blackwell, C.~Alcicek, R.~Fearon, A.~De~Maria,
  V.~Panneershelvam, M.~Suleyman, C.~Beattie, S.~Petersen, et~al.
\newblock Massively parallel methods for deep reinforcement learning.
\newblock \emph{arXiv preprint arXiv:1507.04296}, 2015.

\bibitem[Pfaff et~al.(2020)Pfaff, Fortunato, Sanchez-Gonzalez, and
  Battaglia]{pfaff2020learning}
T.~Pfaff, M.~Fortunato, A.~Sanchez-Gonzalez, and P.~Battaglia.
\newblock Learning mesh-based simulation with graph networks.
\newblock In \emph{International Conference on Learning Representations}, 2020.

\bibitem[Raissi et~al.(2019)Raissi, Perdikaris, and
  Karniadakis]{raissi2019physics}
M.~Raissi, P.~Perdikaris, and G.~E. Karniadakis.
\newblock Physics-informed neural networks: A deep learning framework for
  solving forward and inverse problems involving nonlinear partial differential
  equations.
\newblock \emph{Journal of Computational physics}, 378:\penalty0 686--707,
  2019.

\bibitem[Rolnick et~al.(2019)Rolnick, Ahuja, Schwarz, Lillicrap, and
  Wayne]{rolnick2019experience}
D.~Rolnick, A.~Ahuja, J.~Schwarz, T.~Lillicrap, and G.~Wayne.
\newblock Experience replay for continual learning.
\newblock \emph{Advances in Neural Information Processing Systems}, 32, 2019.

\bibitem[Satorras et~al.(2021)Satorras, Hoogeboom, and Welling]{satorras2021n}
V.~G. Satorras, E.~Hoogeboom, and M.~Welling.
\newblock E (n) equivariant graph neural networks.
\newblock In \emph{International conference on machine learning}, pages
  9323--9332. PMLR, 2021.

\bibitem[Schaul et~al.(2015)Schaul, Quan, Antonoglou, and
  Silver]{schaul2015prioritized}
T.~Schaul, J.~Quan, I.~Antonoglou, and D.~Silver.
\newblock Prioritized experience replay.
\newblock \emph{arXiv preprint arXiv:1511.05952}, 2015.

\bibitem[Stevens et~al.(2020)Stevens, Taylor, Nichols, Maccabe, Yelick, and
  Brown]{stevens2020ai}
R.~Stevens, V.~Taylor, J.~Nichols, A.~B. Maccabe, K.~Yelick, and D.~Brown.
\newblock Ai for science: Report on the department of energy (doe) town halls
  on artificial intelligence (ai) for science.
\newblock Technical report, Argonne National Lab.(ANL), Argonne, IL (United
  States), 2020.

\bibitem[Terraz et~al.(2017)Terraz, Ribes, Fournier, Iooss, and
  Raffin]{terraz2017melissa}
T.~Terraz, A.~Ribes, Y.~Fournier, B.~Iooss, and B.~Raffin.
\newblock Melissa: large scale in transit sensitivity analysis avoiding
  intermediate files.
\newblock In \emph{Proceedings of the international conference for high
  performance computing, networking, storage and analysis}, pages 1--14, 2017.

\bibitem[Um et~al.(2020)Um, Brand, Fei, Holl, and Thuerey]{um2020solver}
K.~Um, R.~Brand, Y.~R. Fei, P.~Holl, and N.~Thuerey.
\newblock Solver-in-the-loop: Learning from differentiable physics to interact
  with iterative pde-solvers.
\newblock \emph{Advances in Neural Information Processing Systems},
  33:\penalty0 6111--6122, 2020.

\bibitem[Vitter(1985)]{vitter1985random}
J.~S. Vitter.
\newblock Random sampling with a reservoir.
\newblock \emph{ACM Transactions on Mathematical Software (TOMS)}, 11\penalty0
  (1):\penalty0 37--57, 1985.

\end{thebibliography}

\end{document}